\documentclass[conference]{IEEEtran}
\usepackage{times}

\usepackage[numbers]{natbib}                                   
\usepackage{algorithm}
\usepackage{algpseudocode}
\usepackage{graphicx} 

\usepackage{tabu}

\usepackage{amsmath}
\usepackage{amssymb}
\usepackage{commath}

\usepackage{color}

\usepackage{multicol}
\usepackage{subcaption}
\usepackage{bm}

\usepackage[bookmarks=true]{hyperref}

\IEEEoverridecommandlockouts 
\title{Leveraging Reward Gradients For Reinforcement Learning in Differentiable Physics Simulations}

\author{Sean Gillen, Katie Byl
\thanks{Sean Gillen and Katie Byl are with the Electrical and Computer Engineering Department at the University of California, Santa Barbara CA 93106
{\tt\small sgillen@ucsb.edu},
{\tt\small katiebyl@ucsb.edu}}
\thanks{Source code is available at: https://github.com/sgillen/apg}%
}

\begin{document}




%

\maketitle


\begin{abstract}
 In recent years, fully differentiable rigid body physics simulators have been developed, which can be used to simulate a wide range of robotic systems. In the context of reinforcement learning for control, these simulators theoretically allow algorithms to be applied directly to analytic gradients of the reward function. However, to date, these gradients have proved extremely challenging to use, and are outclassed by algorithms using no gradient information at all. In this work we present a novel algorithm, cross entropy analytic policy gradients, that is able to leverage these gradients to outperform state of art deep reinforcement learning on a certain set of challenging nonlinear control problems. 

\end{abstract}

\IEEEpeerreviewmaketitle

\section{Introduction}

Computer simulation has become an indispensable tool for researchers and engineers of robotic systems for design, control, and verification. Recent advances in model free deep reinforcement learning (DRL) have been able to leverage simulation and the continued exponential increase computer resources to solve a variety of challenging control and perception problems. Examples include dexterously manipulating objects with a 24 DOF robotic hand~\cite{openai_learning_2018}, and recent work from the Robotic Systems Lab at ETH Zurich demonstrates rough terrain quadruped robot arguably on part with Boston Dynamics  \cite{anymal2022}. In both cases, training was done in simulation, and then successfully transferred to the real world. 

These simulations have largely been treated as black boxes by DRL algorithms. This is both a strength and a weakness of DRL, and this stems partially from the fact that in commonly used simulators like MuJoco \cite{todorov2012mujoco} or Bullet \cite{coumans2020}, are not differentiable (nor are physical robots, for that matter). That is, we are unable to compute the gradient of the state at time t+1, with the action from time t. Therefore we are also unable to take the gradients of the reward function with respect to controller parameters. Thus, RL must thus rely on various approximations of the true gradients, like finite differences or various policy gradient algorithms, the classic example being REINFORCE \cite{williams1992simple}

However in recent years, fully differentiable physics simulators have started to emerge \cite{hu2020difftaichi} \cite{heiden2021neuralsim} \cite{brax2021github}, which offer analytic gradients using automatic differentiation. These simulators already have a number of interesting applications. For example, it is possible to use data from a physical system as data for a learning algorithm to make simulation to better match. Furthermore the nature of some these simulators allow them to be run on hardware accelerators, which offers some obvious speed advantages for algorithms which can make use of them. And, most relevant to this work, they also provide analytic gradients for any differentiable function of simulation state variable. This means we can use stochastic gradient descent, which is the gold standard for training neural networks, directly using the negative sum of rewards as the loss function. We will call training policies in this way an analytic policy gradient algorithm.

However, in practice, these gradients have proven extremely challenging to use, for a number of reasons. Part of the problem is that to take the gradient through any iterated dynamical system required back propagation through time (BPTT). Long chains of BPTT have long been known to cause exploding or vanishing gradients, which naturally causes to difficulty in learning \cite{279181}. A recent paper by Metz et. al. \citet{metz2021gradients} also offer some exposition on the challenges of using the analytic gradients offered by these new rigid body simulators. They highlight that in addition to problems inherent to BPTT, the naturally chaotic dynamics of many rigid body systems exasperate the problems with diverging gradients significantly.

\begin{figure}[!htb]
        \centering
        \includegraphics[width=\linewidth]{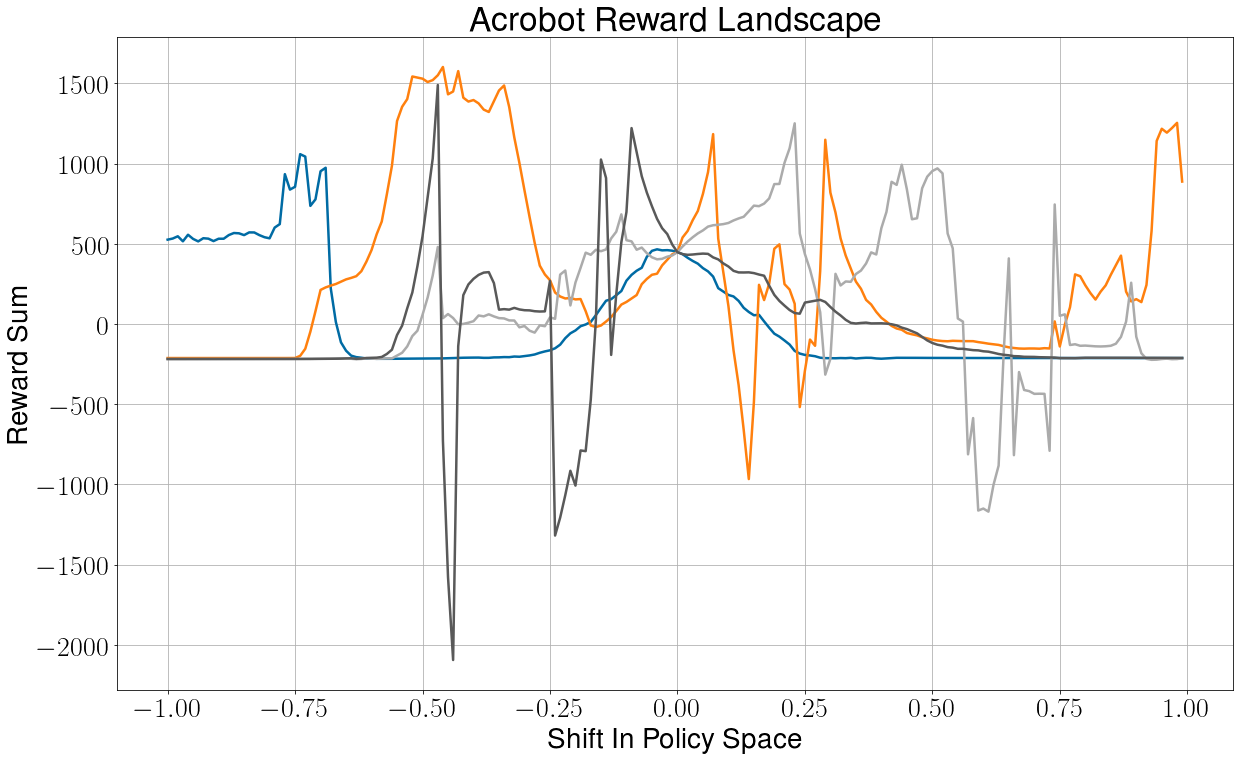}
        \caption{A visualization of the reward landscape for the Acrobot. We started with a random neural network policy, and then sampled a random vector from parameter space. We then added used this vector to shift the initial policy, and record the sum of rewards for a single rollout from the shifted policy . We use the same initial condition for each trial, the only difference between rollouts is that a shifted different policy is used.}
        \label{fig:reward_landscape}
\end{figure}

Another difficulty are severe local maxima. Local maxima are a common problem in all of deep learning, but it is apparent that using APG for reinforcement learning with rigid body systems is especially prone to falling into these extrema. In \cite{metz2021gradients} they also show the reward landscape of the Ant system in Brax. The Ant is a standard benchmark problem in the RL community, it consists of a quadruped robot who's objective is to move forward as fast as possible. They show that reward landscape and reward gradient for this system has extremely high variance, and fraught with local extrema. One may expect that this is due primarily to the fact that the Ant system is relatively high dimensional and subject to contact and frictional forces with the ground. However as we show in figure \ref{fig:reward_landscape}, an Acrobot system, which is a simple two DOF system that is not subject to any contact forces, suffers from many of the same problems. 


Given these difficulties, it remains an open question what role, if any, analytic policy gradients have to play in reinforcement learning for robotic control. In this paper we present a novel algorithm, Cross Entropy Analytic Policy Gradients (CE-APG), which we have found is able to outperform both vanilla APG and state of the art DRL algorithms in at least one class of challenging nonlinear control problems. Specifically, we demonstrate that under-actuated planar kinematic chains, like the acrobat or inverted cart pole pendulum can be successfully controlled by this algorithm.

CE-APG uses APG as a local search, combining it with an outer loop cross entropy method to escape from local maxima in the reward landscape. This was inspired the the approach taken by the authors of Tiny Differentiable Simulator \cite{heiden2021neuralsim}, which used an optimization technique called Parallel Basin Hopping combined with a gradient based algorithm for what was essentially system ID.

\section{Background And Related Work}
\subsection{Deep Reinforcement Learning}

Deep Reinforcement Learning has seen a lot of attention and impressive results over the last decade or so, including in the context of continuous control for robotic systems \cite{heess_emergence_2017} \cite{openai_learning_2018} \cite{lee_robust_2019} \cite{siekmann2021blind}. These problems are all high dimensional, nonlinear, underactuated, and they all involve complex contact sequences with the environments, which makes them very challenging for more traditional control design. 

DRL is usually divided into model based and model free control. Model based reinforcement learning learns a model of the system under control, and uses that to do planning, classic dynamic programming is an example, as is PILCO \cite{deisenroth2011pilco}. These approaches are typically much more sample efficient than model free RL, however these methods typically have a hard time scaling to higher dimensional problems, and can require expensive re planning at run time.

Model free RL on the other hand learns a policy directly to maximize a reward function. This implies solving a more difficult optimization problem, but is what has been used to achieve all the results we have discussed in this paper. Examples of model free algorithms include Soft Actor Critic (SAC), Proximal Policy Optimization (PPO), and Twin Delayed Deep Deterministic policy gradient (TD3) \cite{haarnoja2018soft} \cite{schulman2017proximal} \cite{fujimoto2018addressing}.

Although they do not strictly use nueral networks, gradient free methods like Evolutionary Strategies (ES) \cite{salimans2017evolution}, or Augmented Random Search (ARS) \cite{Mania2018} can also be considered modern model free reinforcement learning algorithms.  

Our algorithm can be considered a model free reinforcement learning algorithm, and inherits many of their advantages and disadvantages. For example the algorithm we present should theoretically scale well to high dimensional tasks. However our method currently has sample efficiency on par with other DRL algorithms, and, as it requires a differentiable simulator to function, will require sim2real transfer to be applicable to physical systems. We do note that successful transfer of polices from simulation directly to hardware has been demonstrated many times in the literature \cite{openai_learning_2018} \cite{anymal2022}. 


\subsection{Policy Search in Differentiable Simulations}

Many of the papers which introduce a differentiable simulator also include a basic example using analytic gradients for policy search. Brax \cite{brax2021github} and the unnamed simulator developed by Degrave et. al. \cite{10.3389/fnbot.2019.00006} both implement a basic version of APG. Brax's APG is used to command a fully actuated double pendulum to reach random targets, but is currently unable to solve most of the other problems in their benchmarking suite. Degrave Et. Al. manage to develop a walking gait for a quadruped, though their method requires a fairly significant amount of hand designed components.

In \cite{qiao2021efficient} They introduce a simulator and suggest something called "policy enhancement" whereby they augment a model based RL algorithm with the analytic gradients to control a fully actuated double pendulum.In \cite{pmlr-v139-mora21a} the authors present policy optimization via differentiable simulation. They don't directly use the analytic policy gradient, and instead develop an indirect second order method. They also demonstrate their system on under-actuated systems like the inverted pendulum, the difference is that they are balancing at the stable equilibrium. This is a deceptively difficult problem, however we show that our method works to swing-up and balance the system in their unstable equilibrium, which we would argue is more difficult.

\subsection{Combining Local and Global Search}

As already mentioned, our algorithm was inspired by basin hopping \cite{doi:10.1021/jp970984n} and the extension of parallel basin hopping \cite{doi:10.2514/6.2018-1452}. Tiny Differentiable Simulator \cite{heiden2021neuralsim} uses this method for parameter estimation in their own work to perform system ID. We instead use a cross entropy method, and are obviously tackling a different method.

There are also several methods that combine a zeroth order optimizer with a local gradient based optimizer for robot learning \cite{pourchot2019cemrl} \cite{bharadhwaj2020modelpredictive} \cite{huang2021cemgd} \cite{pourchot2019cemrl}. However none of them are making use of analytic gradients, or doing direct policy optimization.







 
  

\subsection{Back Propagation Through Time}



In order to propagate gradients through an iterated system, we must use a technique called back propagation through time. As we have mentioned, this causes difficulty in exploding or vanishing gradients, especially in long chains of computation. The problem essentially is that the reward at time t depends not only on the state and action at time t-1, but on the state and action from t-2, t-3, back to the initial state, even if system itself is Markovian. For even modest length roll-outs this causes instability in the gradients that make learning with them challenging. 

There are many other contexts that this problem arises, including in natural language processing (NLP). One of the tools used to combat this in the NLP community are specialized recurrent neural networks, which are specifically designed to stop gradients that pass through the network from diverging. In our case we use gated recurrent unit GRU \cite{8053243} as our control policy, we outline this architecture with more detail in the methods section. 



\section{Problem Formulation}

\subsection{Reinforcement Learning}

In reinforcement learning, the goal is to train an agent, acting in an environment, to maximize a scalar reward function. The environment is a discrete time dynamical system described by state $s_{t} \in \mathbb{R}^{n}$ and the current action $a_{t} \in \mathbb{R}^{b}$. An evolution function $f: \mathbb{R}^{n} \times \mathbb{R}^{b} \rightarrow \mathbb{R}^{n}$ takes as input the current state and action, and outputs the state at time t+1:

\begin{equation}
s_{t+1}= f(s_{t},a_{t})
\end{equation} 

The controller is a function parameterized by a vector $\theta \in \mathbb{R}^{\abs{\theta}}$ that maps states to actions $g: \mathbb{R}^{n} \times \mathbb{R}^{\norm{\theta}} \rightarrow \mathbb{R}^{m}$ such that:

\begin{equation}
a_{t} = g(s_{t}, \theta)
\end{equation} 

The goal is to maximize a scalar reward function $r : \mathbb{R}^{n} \times \mathbb{R}^{m} \times \mathbb{R}^{n} \rightarrow \mathbb{R}$. We consider the finite time undiscounted reward case. The objective function then is:

\begin{equation} 
R(\theta) =  \sum_{t=0}^{T}r(s_{t}, a_{t}, s_{t+1}) 
\end{equation}




\subsection{Kinematic Chains and Simulation}

We consider three systems, a classic cartpole pendulum, a double cartpole pendulum, and an Acrobot \cite{spong_swing_1994}. These are all under-actuated, kinematic chains, and are often used as benchmark problems for both reinforcement learning and nonlinear control. Their dynamics in general can be described with the following:

\begin{equation}
    \textbf{M}(\textbf{q})\Ddot{\textbf{q}} + \textbf{C}(\textbf{q},\dot{\textbf{q}})\dot{\textbf{q}} + T_g(\textbf{q})g = \textbf{B}\textbf{u} \label{manipEqn}
\end{equation}

with M being the inertia matrix, C being the Coriolis matrix, T being a matrix capturing gravitational affects. For these systems u is the torque outputted at each motor, and q is the vector of state variables. In each of these systems, there is a single unstable equilibrium point. Our goal is to swing the system from an initial condition to it's unstable point and the maintain balance there. More details on formulating these environments into an MDP can be found in the appendix. 

It was demonstrated in \cite{Gillen2020CombiningDR} that most DRL struggles with the full version of the Acrobot in particular. It is worth elaborating on that point. Many benchmarking suites (for example, OpenAI's gym \cite{1606.01540}) have underactuated systems like the acrobot or cartpole pendulum, however the tasks are typically either to swing the system up, or to balance it, asking one controller to do both becomes a much more challenging problem. 


\subsection{The Brax Simulator}

The above systems are simulated using Brax \cite{brax2021github}. Brax is a differentiable physics engine that can simulate systems made up of rigid bodies, joint constraints, and actuators. The simulators primary advantage is that it can run massively parallel simulations very quickly on accelerator hardware, I.E. TPUs and GPUs. By virtue of being written entirely in Jax \cite{jax2018github}, we can also take arbitrary gradients through the simulator using autodiff. 



\section{Methods}

\subsection{Analytic policy gradients}

Stochastic gradient descent and its variants (in our case adam \cite{kingma2017adam}) are the gold standard for training deep neural networks. We seek to train our policy using the gradient of the sum of the reward function for a given episode. 

\begin{equation}
    \label{eq:vapg}
    \theta^{+} = \theta + \alpha \nabla_\theta R(\theta)
\end{equation}

To be more specific, we perform N policy rollouts using the current parameters, take the gradient of the mean of the sum of rewards for each these roll outs, use that gradient to update the current policy, and repeat until convergence. Thus our update step is:

\begin{equation}
    \label{eq:vapg_sum}
    \theta^{+} = \theta + \alpha \nabla_\theta \frac{1}{N}\sum_{i=0}^{N}\sum_{t=0}^{T}r(s_{t}, a_{t}, s_{t+1}) 
\end{equation}

As we've discussed, because we are using a differentiable simulation, the analytic gradient with respect to $\theta$ is available to us.




\subsection{The Cross Entropy Method}

The Cross Entropy Method (CEM \cite{RUBINSTEIN199789}) is a well established algorithm for importance sampling and optimization. CEM maintains a probability distribution over its decision variables, in this case the decision variables are the parameters for our policy. The most common formulation is to use a normal Distribution, thus we must mantain a vector of means $\mu_{pi}$, and a covariance matrix $\sigma_{\pi}$. $\mathcal{N}(\mu_{\pi}, \sigma_{\pi})$. At each step we sample candidate policies from this distribution, and use the following update rules: 

\begin{equation}
\label{eq:mean}
    \mu_{\pi}^{+} = \frac{1}{K_{e}}\sum_{i=0}^{K_{e}}\mu_{i}
\end{equation}

\begin{equation}
\label{eq:var1}
    \sigma_{\pi}^{2+} = \frac{1}{K_{e}}\sum_{i=0}^{K_{e}}(\mu_{\pi} - \mu_{i})(\mu_{\pi} - \mu_{i})^{T}
\end{equation}

Howeve, the covariance matrix grows quadratically with the number of policy parameters, and neural networks can have thousands of parameters even for small systems. Thus, we make the following simplification to the variance:

\begin{equation}
\label{eq:var}
    \sigma_{\pi}^{2+} = \frac{1}{K_{e}}\sum_{i=0}^{K_{e}}(\mu_{\pi} - \mu_{i})^{2}
\end{equation}

This implicitly ensures that our covariance "matrix" only has entries on the diagonal, and can thus be stored as a vector. This simplification is also made by \cite{pourchot2019cemrl}.

\subsection{Cross Entropy Analytic Policy Gradients}

\begin{algorithm}
\caption{Cross Entropy Analytic Policy Gradients}
\label{algo:ceapg}
\begin{algorithmic}
\Require Policy $\pi$ with trainable parameters $\theta$
\Require Hyper-parameters - $\sigma_{0}$, $K_{a}$, $K_{e}$
\State Sample $\bm{\theta_{c}} = [\theta_{1}, ..., \theta_{n}]$ from $\mathcal{N}(\theta, \sigma^{2})^{K_{a} \times \abs{\theta}}$
\For{$\theta_{i} \text{ in } \bm{\theta_{c}}$}
    \State Run APG with initial policy weights $\theta_{i}$
    \State Collect sum of rewards $R_{i}$ and final policy $\theta^{*}_{i}$. 
\EndFor
\State Sort $\theta^{*}$ values in descending order according to reward
\State $\theta^{+} = \frac{1}{K_{e}}\sum_{i=0}^{K_{e}}\theta^{*}_{i} $
\State $\sigma^{+} = \sqrt{\frac{1}{K_{e}}\sum_{i=0}^{K_{e}}(\theta - \theta^{*}_{i})^{2}}$

\end{algorithmic}
\end{algorithm}

We combine these two algorithms as follows. Start with initial policy weight $\theta$, and an initial parameter variance $\sigma_{0}$. We then generate $K_{a}$ candidate policies by sampling from $\mathcal{N}(\theta, \sigma_{0})$. Using these policies as initial conditions, we run $K_{a}$ analytic policy gradient algorithms in parallel, which gives us new weight vector $ [ \theta_{0} , \theta_{1} ... \theta_{K_{a}} ]$, and the final rewards for these new policy weight, $R_{1}, R_{2} ... R_{K_{a}}$. We then sort the policy weights in descending order based on their associated final return. Finally we select the top $K_{e}$ and use equations \ref{eq:mean} and \ref{eq:var} to update our parameter and variance vector. This is repeated until some stopping criteria, for this paper we simply train for fixed number of steps.

\subsection{Controller Architecture}

As we have already mentioned, we employ a Gated Recurrent Unit (GRU) network as our control policy to help combat the exploding / vanishing gradient problem. For our CE-APG experiments, we used a GRU with two fully connected layers on the output, with ReLU hidden activations and a Tanh non-linearity on the final layer. Network sizes for each experiment can be found in the appendix. We found that by using the GRU we are able to train with episodes lengths of at least 500 steps. 


In addition to this, we use deterministic policies, rather than the stochastic ones usually associated with deep reinforcement learning. Typically in DRL, the policy actually parameters a probability distribution over possible actions. At every time step, one generates a new distribution based on the current state, and then samples from that distribution to select an action. While our algorithm is compatible with stochastic policies of this nature, we instead compute the action directly. We believe this is especially advantageous for systems with unstable and highly sensitive dynamics (like the acrobot, cartpole etc).

\section{Results}

\begin{figure*}[!h]
  \centering
  \begin{subfigure}[b]{0.32\linewidth}
    \includegraphics[width=\linewidth]{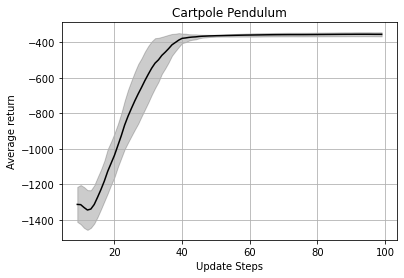}
  \end{subfigure}
  \begin{subfigure}[b]{0.32\linewidth}
  \includegraphics[width=\linewidth]{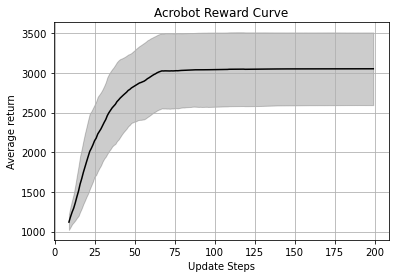}
  \end{subfigure}
  \begin{subfigure}[b]{0.32\linewidth}
      \includegraphics[width=\linewidth]{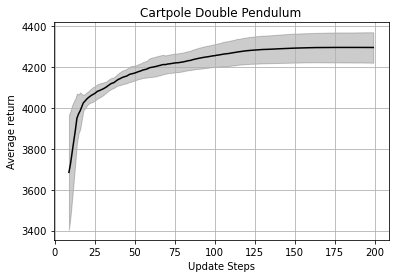}
  \end{subfigure}
  \caption{Reward curves for CE-APG}
  \label{fig:reward_curves}
\end{figure*}

For each environment, we ran trials with 8 random seeds. The seeds affect all the sources of randomness during training, of which there are several. The initial value of the policy parameters, the noise added to the policy at the beginning of each iteration, and the initial condition of the simulator at the beginning of each episode. Figure \ref{fig:reward_curves}. For each of these trials we used the GRU controller architecture discussed above, details on layer sizes etc. can be found in the appendix. We report the resulting rewards obtained at the end of in table \ref{tab:results1}. In addition to our own algorithm, we also run comparisons from PPO, SAC, and Brax's implementation of APG (which has some note-able difference from our own, discussed in the appendix).

\begin{table}[!htb]
\centering
\renewcommand{\arraystretch}{1.5}
\begin{tabular}{|l|l|l|l|}
\hline 
Environment   & Pendulum              & Acrobot               & Double Pendulum \\ \hline \hline
CE-APG (ours)    &   -355$\pm$ 10  &   3053 $\pm$ 458       &   4295 $\pm$ 75\\  \hline
PPO               &    -464 $\pm$ 223 &    405 $\pm$ 927       &   4249 $\pm$ 549         \\  \hline
SAC               &    -2274  $\pm$ 1000    &    359 $\pm$ 159      &   -3.9e5 $\pm$ 4.9e6        \\ \hline
Brax Apg        &    -3485   $\pm$ 819    &     1949 $\pm$ 814       &   -1982 $\pm$ 4056     \\ 

\hline
\end{tabular}
\caption{Results of the training on our test environments, we report the mean and standard deviation of rewards obtained from training each algorithm with 8 random seeds}
\label{tab:results1}
\end{table}

We can see that across all three environments, our method outperforms the other benchmark algorithms. On the double inverted pendulum we get the same final reward as PPO, exceed it slightly on the cartpole, and get significantly higher reward on the acrobot. In fact for the cart-pole in particular our algorithm significantly outperforms the others. It is worth putting this in context, as it is difficult to understand what a reward of 2000 vs. 3000 really means.

\begin{figure}[!htb]
        \centering
        \includegraphics[width=\linewidth]{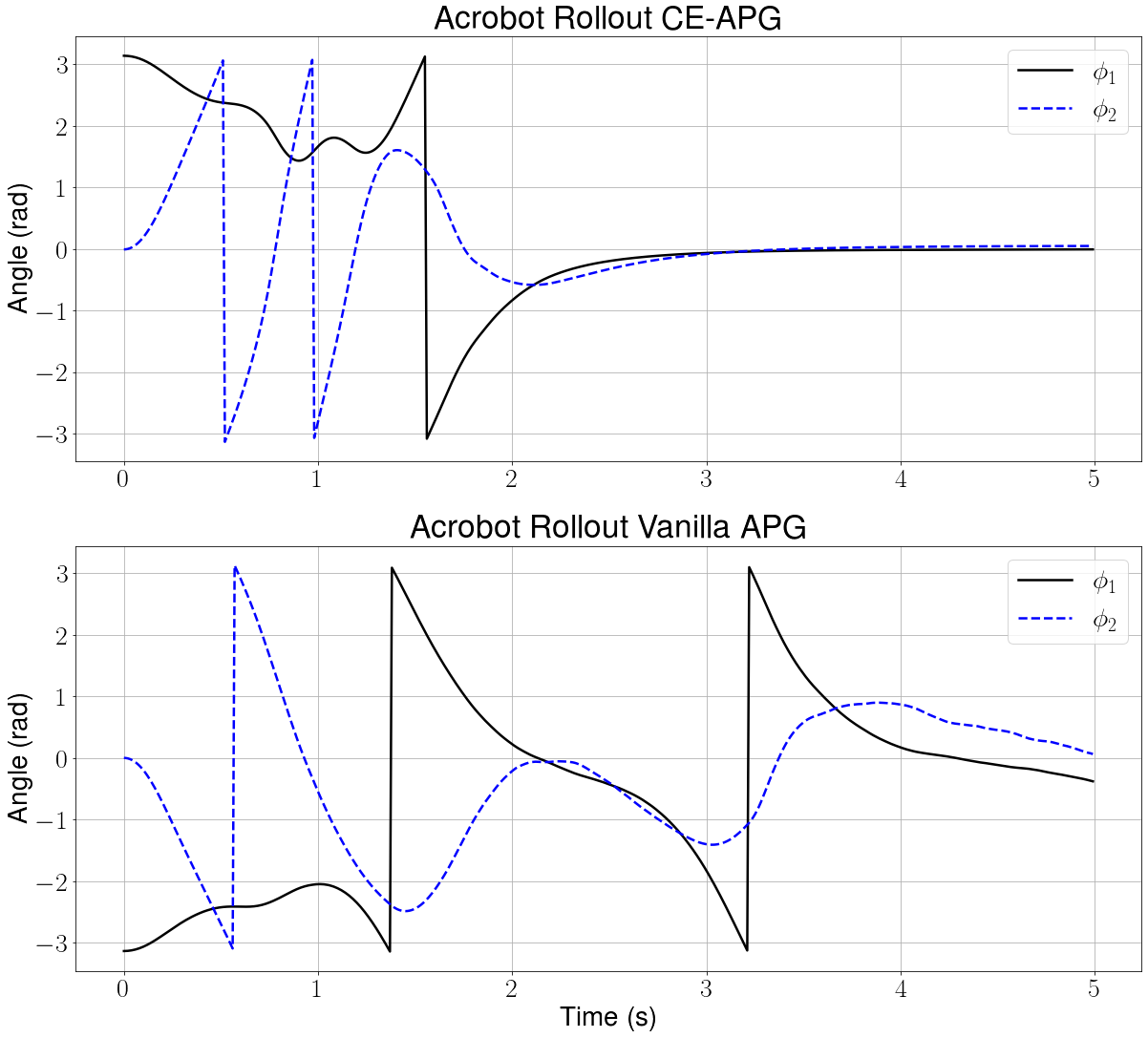}
        \caption{Best performing seeds for CE-APG vs APG on the Acrobot. Despite comparable total reward, APG does not actually stabilize the system at the equilibrium, likely because it has become stuck in a maxima of the reward landscape. By contrast our algorithm, augmented by the cross entropy method to avoid such local maxima, manages to find a policy that does stabalize our system}
        \label{fig:rollouts}
\end{figure}

To do this we perform roll-outs with the best performing seeds for both  CE-APG and the best performing benchmark, which was Brax's APG implementation. We perform 10 rollouts with these top performers. For completeness we report the mean and standard deviation of the resulting rewards, CE-APG: 3507 $\pm$ 5, Brax's APG: 2778 $\pm$ 89. However it is much more revealing to visualize one of these roll outs, which we do in figure \ref{fig:rollouts}. We can see that despite comparable total reward, APG does not actually stabilize the system at the equilibrium, likely because it has become stuck in a maxima of the reward landscape. By contrast, our algorithm, augmented by the cross entropy method to avoid such local maxima, manages to find a policy that does stabilize our system. Unsurprisingly given the rewards from table \ref{tab:results1} none of the other systems manage to balance the acrobat either. Of course such stabilization is exactly what we as humans had in mind when defining the environment. 

\section{Ablation Studies}

In addition to the comparisons done above, we also conducted several ablation studies, this isolates the effect of our implementation of APG from the results presented above. We conducted experiments using only the gradient free CEM method, with no analytic policy gradients being used. And also compared to our implementation of APG, which we are calling PAPG for parallel APG. This essentially means that we run APG N times in parallel, and picking the best result to return. In both cases we still used the GRU controller architecture as before. In the case of CEM we used 2000 iterations for every environment, which we found to be well past the point of convergence. For APG we used 200*100 iterations, which results in the same number of environment interactions to CE-APG. The results are shown in table \ref{tab:ablation}.

\begin{table}[!htb]
\centering
\renewcommand{\arraystretch}{1.5}
\begin{tabular}{|l|l|l|l|}
\hline 
Environment         & Pendulum            & Acrobot & Double Pendulum \\ \hline \hline
CE-APG (ours)    &   -355$\pm$ 10         &   3053 $\pm$ 458       &   4295 $\pm$ 75\\  \hline
PAPG             &    -1034 $\pm$ 208     &   2456 $\pm$ 647       &  2335 $\pm$ 431  \\ \hline
CEM              &    -3019 $\pm$ 559     &   626 $\pm$ 160        &  -2.7e5 $\pm$ 7.6e5  \\ \hline

\hline
\end{tabular}
\caption{Results of our ablation studies. We report the mean and standard deviation of rewards obtained from training each algorithm with 8 random seeds}
\label{tab:ablation}
\end{table}

As we can see, the performance of either method individually is generally poor, though we do see that PAPG does about as well as the best agent from the APG results reported in table \ref{tab:results1}, however when we perform rollouts the resulting policies exhibit the same behavior shown in figure \ref{fig:rollouts}, that is they continuously spin, spending as much time as they can near the goal state, but never settling there.

\section{Discussion and future work}

    We have shown that analytic policy gradients can be leveraged effectively for at least one class of system, nonlinear underacted systems with unstable target states. This is a limited scope of problems, but it is interesting because other modern DRL algorithms struggle with such systems. However, we are as of yet unable to get our algorithm to perform well in contact rich environments, likely because the contacts introduce huge variance into the gradients. As of writing, there is active work in the community to make Brax friendlier to analytic gradient based algorithms, in particular adding soft contacts, which may very well help a lot. 
    
    In addition, the sample complexity of our algorithm is comparable to other on policy DRL algorithm, and is behind what we might expect from off policy algorithms. However there are many algorithmic improvements could be made, for example importance sampling has been found to improve the sample efficiency of CEM by up to a factor of 10. We are currently unable to effectively implement this due technical limitations in the way we implement parallelization. 
\section{Conclusion} 
\label{sec:conclusion}

In conclusion, we presented Cross Entropy Analytic Policy Gradients, an algorithm that can exploit analytic gradients. We covered some relevant background, introduced our method, and placed it in the broader context. We then presented our algorithm and the environments we used for testing. We presented results of our algorithm compared to state of the art baselines, and performed ablation studies. We then contextualized the rewards obtained on Acrobat in particular. This demonstrated that our algorithm was able to successfully stabilize the system, whereas the baseline algorithms where not. We think this algorithm shows that analytic policy gradients can be leveraged productively in at least context.

\bibliographystyle{plainnat}
\bibliography{main}

\begin{thebibliography}{34}
\providecommand{\natexlab}[1]{#1}
\providecommand{\url}[1]{\texttt{#1}}
\expandafter\ifx\csname urlstyle\endcsname\relax
  \providecommand{\doi}[1]{doi: #1}\else
  \providecommand{\doi}{doi: \begingroup \urlstyle{rm}\Url}\fi

\bibitem[Bengio et~al.(1994)Bengio, Simard, and Frasconi]{279181}
Y.~Bengio, P.~Simard, and P.~Frasconi.
\newblock Learning long-term dependencies with gradient descent is difficult.
\newblock \emph{IEEE Transactions on Neural Networks}, 5\penalty0 (2):\penalty0
  157--166, 1994.
\newblock \doi{10.1109/72.279181}.

\bibitem[Bharadhwaj et~al.(2020)Bharadhwaj, Xie, and
  Shkurti]{bharadhwaj2020modelpredictive}
Homanga Bharadhwaj, Kevin Xie, and Florian Shkurti.
\newblock Model-predictive control via cross-entropy and gradient-based
  optimization, 2020.

\bibitem[Bradbury et~al.(2018)Bradbury, Frostig, Hawkins, Johnson, Leary,
  Maclaurin, Necula, Paszke, Vander{P}las, Wanderman-{M}ilne, and
  Zhang]{jax2018github}
James Bradbury, Roy Frostig, Peter Hawkins, Matthew~James Johnson, Chris Leary,
  Dougal Maclaurin, George Necula, Adam Paszke, Jake Vander{P}las, Skye
  Wanderman-{M}ilne, and Qiao Zhang.
\newblock {JAX}: composable transformations of {P}ython+{N}um{P}y programs,
  2018.
\newblock URL \url{http://github.com/google/jax}.

\bibitem[Brockman et~al.(2016)Brockman, Cheung, Pettersson, Schneider,
  Schulman, Tang, and Zaremba]{1606.01540}
Greg Brockman, Vicki Cheung, Ludwig Pettersson, Jonas Schneider, John Schulman,
  Jie Tang, and Wojciech Zaremba.
\newblock Openai gym, 2016.

\bibitem[Coumans and Bai(2016--2020)]{coumans2020}
Erwin Coumans and Yunfei Bai.
\newblock Pybullet, a python module for physics simulation for games, robotics
  and machine learning.
\newblock \url{http://pybullet.org}, 2016--2020.

\bibitem[Degrave et~al.(2019)Degrave, Hermans, Dambre, and
  wyffels]{10.3389/fnbot.2019.00006}
Jonas Degrave, Michiel Hermans, Joni Dambre, and Francis wyffels.
\newblock A differentiable physics engine for deep learning in robotics.
\newblock \emph{Frontiers in Neurorobotics}, 13, 2019.
\newblock ISSN 1662-5218.
\newblock \doi{10.3389/fnbot.2019.00006}.
\newblock URL
  \url{https://www.frontiersin.org/article/10.3389/fnbot.2019.00006}.

\bibitem[Deisenroth and Rasmussen(2011)]{deisenroth2011pilco}
Marc Deisenroth and Carl~E Rasmussen.
\newblock Pilco: A model-based and data-efficient approach to policy search.
\newblock In \emph{Proceedings of the 28th International Conference on machine
  learning (ICML-11)}, pages 465--472. Citeseer, 2011.

\bibitem[Dey and Salem(2017)]{8053243}
Rahul Dey and Fathi~M. Salem.
\newblock Gate-variants of gated recurrent unit (gru) neural networks.
\newblock In \emph{2017 IEEE 60th International Midwest Symposium on Circuits
  and Systems (MWSCAS)}, pages 1597--1600, 2017.
\newblock \doi{10.1109/MWSCAS.2017.8053243}.

\bibitem[Freeman et~al.(2021)Freeman, Frey, Raichuk, Girgin, Mordatch, and
  Bachem]{brax2021github}
C.~Daniel Freeman, Erik Frey, Anton Raichuk, Sertan Girgin, Igor Mordatch, and
  Olivier Bachem.
\newblock Brax - a differentiable physics engine for large scale rigid body
  simulation, 2021.
\newblock URL \url{http://github.com/google/brax}.

\bibitem[Fujimoto et~al.(2018)Fujimoto, van Hoof, and
  Meger]{fujimoto2018addressing}
Scott Fujimoto, Herke van Hoof, and David Meger.
\newblock Addressing function approximation error in actor-critic methods,
  2018.

\bibitem[Gillen et~al.(2020)Gillen, Molnar, and Byl]{Gillen2020CombiningDR}
S.~Gillen, Marco Molnar, and K.~Byl.
\newblock Combining deep reinforcement learning and local control for the
  acrobot swing-up and balance task.
\newblock \emph{2020 59th IEEE Conference on Decision and Control (CDC)}, pages
  4129--4134, 2020.

\bibitem[Haarnoja et~al.(2018)Haarnoja, Zhou, Abbeel, and
  Levine]{haarnoja2018soft}
Tuomas Haarnoja, Aurick Zhou, Pieter Abbeel, and Sergey Levine.
\newblock Soft actor-critic: Off-policy maximum entropy deep reinforcement
  learning with a stochastic actor, 2018.

\bibitem[Heess et~al.(2017)Heess, TB, Sriram, Lemmon, Merel, Wayne, Tassa,
  Erez, Wang, Eslami, Riedmiller, and Silver]{heess_emergence_2017}
Nicolas Heess, Dhruva TB, Srinivasan Sriram, Jay Lemmon, Josh Merel, Greg
  Wayne, Yuval Tassa, Tom Erez, Ziyu Wang, S.~M.~Ali Eslami, Martin Riedmiller,
  and David Silver.
\newblock Emergence of {Locomotion} {Behaviours} in {Rich} {Environments}.
\newblock \emph{arXiv:1707.02286 [cs]}, July 2017.
\newblock URL \url{http://arxiv.org/abs/1707.02286}.
\newblock arXiv: 1707.02286.

\bibitem[Heiden et~al.(2021)Heiden, Millard, Coumans, Sheng, and
  Sukhatme]{heiden2021neuralsim}
Eric Heiden, David Millard, Erwin Coumans, Yizhou Sheng, and Gaurav~S Sukhatme.
\newblock Neural{S}im: Augmenting differentiable simulators with neural
  networks.
\newblock In \emph{Proceedings of the IEEE International Conference on Robotics
  and Automation (ICRA)}, 2021.
\newblock URL
  \url{https://github.com/google-research/tiny-differentiable-simulator}.

\bibitem[Hu et~al.(2020)Hu, Anderson, Li, Sun, Carr, Ragan-Kelley, and
  Durand]{hu2020difftaichi}
Yuanming Hu, Luke Anderson, Tzu-Mao Li, Qi~Sun, Nathan Carr, Jonathan
  Ragan-Kelley, and Frédo Durand.
\newblock Difftaichi: Differentiable programming for physical simulation, 2020.

\bibitem[Huang et~al.(2021)Huang, Lale, Rosolia, Shi, and
  Anandkumar]{huang2021cemgd}
Kevin Huang, Sahin Lale, Ugo Rosolia, Yuanyuan Shi, and Anima Anandkumar.
\newblock Cem-gd: Cross-entropy method with gradient descent planner for
  model-based reinforcement learning, 2021.

\bibitem[Kingma and Ba(2017)]{kingma2017adam}
Diederik~P. Kingma and Jimmy Ba.
\newblock Adam: A method for stochastic optimization, 2017.

\bibitem[Lee et~al.(2019)Lee, Hwangbo, and Hutter]{lee_robust_2019}
Joonho Lee, Jemin Hwangbo, and Marco Hutter.
\newblock Robust {Recovery} {Controller} for a {Quadrupedal} {Robot} using
  {Deep} {Reinforcement} {Learning}.
\newblock \emph{arXiv:1901.07517 [cs]}, January 2019.
\newblock URL \url{http://arxiv.org/abs/1901.07517}.
\newblock arXiv: 1901.07517.

\bibitem[Mania et~al.(2018)Mania, Guy, and Recht]{Mania2018}
Horia Mania, Aurelia Guy, and Benjamin Recht.
\newblock {Simple random search of static linear policies is competitive for
  reinforcement learning}.
\newblock \emph{Advances in Neural Information Processing Systems},
  2018-December\penalty0 (NeurIPS):\penalty0 1800--1809, 2018.
\newblock ISSN 10495258.

\bibitem[McCarty et~al.()McCarty, Burke, and McGuire]{doi:10.2514/6.2018-1452}
Steven~L. McCarty, Laura~M. Burke, and Melissa McGuire.
\newblock \emph{Parallel Monotonic Basin Hopping for Low Thrust Trajectory
  Optimization}.
\newblock \doi{10.2514/6.2018-1452}.
\newblock URL \url{https://arc.aiaa.org/doi/abs/10.2514/6.2018-1452}.

\bibitem[Metz et~al.(2021)Metz, Freeman, Schoenholz, and
  Kachman]{metz2021gradients}
Luke Metz, C~Daniel Freeman, Samuel~S Schoenholz, and Tal Kachman.
\newblock Gradients are not all you need.
\newblock \emph{arXiv preprint arXiv:2111.05803}, 2021.

\bibitem[Miki et~al.(2022)Miki, Lee, Hwangbo, Wellhausen, Koltun, and
  Hutter]{anymal2022}
Takahiro Miki, Joonho Lee, Jemin Hwangbo, Lorenz Wellhausen, Vladlen Koltun,
  and Marco Hutter.
\newblock Learning robust perceptive locomotion for quadrupedal robots in the
  wild.
\newblock \emph{Science Robotics}, 7\penalty0 (62), Jan 2022.
\newblock ISSN 2470-9476.
\newblock \doi{10.1126/scirobotics.abk2822}.
\newblock URL \url{http://dx.doi.org/10.1126/scirobotics.abk2822}.

\bibitem[Mora et~al.(2021)Mora, Peychev, Ha, Vechev, and
  Coros]{pmlr-v139-mora21a}
Miguel Angel~Zamora Mora, Momchil Peychev, Sehoon Ha, Martin Vechev, and
  Stelian Coros.
\newblock Pods: Policy optimization via differentiable simulation.
\newblock In Marina Meila and Tong Zhang, editors, \emph{Proceedings of the
  38th International Conference on Machine Learning}, volume 139 of
  \emph{Proceedings of Machine Learning Research}, pages 7805--7817. PMLR,
  18--24 Jul 2021.
\newblock URL \url{https://proceedings.mlr.press/v139/mora21a.html}.

\bibitem[OpenAI et~al.(2018)OpenAI, Andrychowicz, Baker, Chociej, Jozefowicz,
  McGrew, Pachocki, Petron, Plappert, Powell, Ray, Schneider, Sidor, Tobin,
  Welinder, Weng, and Zaremba]{openai_learning_2018}
OpenAI, Marcin Andrychowicz, Bowen Baker, Maciek Chociej, Rafal Jozefowicz, Bob
  McGrew, Jakub Pachocki, Arthur Petron, Matthias Plappert, Glenn Powell, Alex
  Ray, Jonas Schneider, Szymon Sidor, Josh Tobin, Peter Welinder, Lilian Weng,
  and Wojciech Zaremba.
\newblock Learning {Dexterous} {In}-{Hand} {Manipulation}.
\newblock \emph{arXiv:1808.00177 [cs, stat]}, August 2018.
\newblock URL \url{http://arxiv.org/abs/1808.00177}.
\newblock arXiv: 1808.00177.

\bibitem[Pourchot and Sigaud(2019)]{pourchot2019cemrl}
Aloïs Pourchot and Olivier Sigaud.
\newblock Cem-rl: Combining evolutionary and gradient-based methods for policy
  search, 2019.

\bibitem[Qiao et~al.(2021)Qiao, Liang, Koltun, and Lin]{qiao2021efficient}
Yi-Ling Qiao, Junbang Liang, Vladlen Koltun, and Ming~C. Lin.
\newblock Efficient differentiable simulation of articulated bodies, 2021.

\bibitem[Rubinstein(1997)]{RUBINSTEIN199789}
Reuven~Y. Rubinstein.
\newblock Optimization of computer simulation models with rare events.
\newblock \emph{European Journal of Operational Research}, 99\penalty0
  (1):\penalty0 89--112, 1997.
\newblock ISSN 0377-2217.
\newblock \doi{https://doi.org/10.1016/S0377-2217(96)00385-2}.
\newblock URL
  \url{https://www.sciencedirect.com/science/article/pii/S0377221796003852}.

\bibitem[Salimans et~al.(2017)Salimans, Ho, Chen, Sidor, and
  Sutskever]{salimans2017evolution}
Tim Salimans, Jonathan Ho, Xi~Chen, Szymon Sidor, and Ilya Sutskever.
\newblock Evolution strategies as a scalable alternative to reinforcement
  learning, 2017.

\bibitem[Schulman et~al.(2017)Schulman, Wolski, Dhariwal, Radford, and
  Klimov]{schulman2017proximal}
John Schulman, Filip Wolski, Prafulla Dhariwal, Alec Radford, and Oleg Klimov.
\newblock Proximal policy optimization algorithms, 2017.

\bibitem[Siekmann et~al.(2021)Siekmann, Green, Warila, Fern, and
  Hurst]{siekmann2021blind}
Jonah Siekmann, Kevin Green, John Warila, Alan Fern, and Jonathan Hurst.
\newblock Blind bipedal stair traversal via sim-to-real reinforcement learning,
  2021.

\bibitem[Spong(1994)]{spong_swing_1994}
Mark~W. Spong.
\newblock Swing up control of the acrobot using partial feedback linearization
  *.
\newblock \emph{IFAC Proceedings Volumes}, 27\penalty0 (14):\penalty0 833--838,
  September 1994.
\newblock ISSN 14746670.
\newblock \doi{10.1016/S1474-6670(17)47404-0}.
\newblock URL
  \url{https://linkinghub.elsevier.com/retrieve/pii/S1474667017474040}.

\bibitem[Todorov et~al.(2012)Todorov, Erez, and Tassa]{todorov2012mujoco}
Emanuel Todorov, Tom Erez, and Yuval Tassa.
\newblock Mujoco: A physics engine for model-based control.
\newblock In \emph{2012 IEEE/RSJ International Conference on Intelligent Robots
  and Systems}, pages 5026--5033. IEEE, 2012.

\bibitem[Wales and Doye(1997)]{doi:10.1021/jp970984n}
David~J. Wales and Jonathan P.~K. Doye.
\newblock Global optimization by basin-hopping and the lowest energy structures
  of lennard-jones clusters containing up to 110 atoms.
\newblock \emph{The Journal of Physical Chemistry A}, 101\penalty0
  (28):\penalty0 5111--5116, 1997.
\newblock \doi{10.1021/jp970984n}.

\bibitem[Williams(1992)]{williams1992simple}
Ronald~J Williams.
\newblock Simple statistical gradient-following algorithms for connectionist
  reinforcement learning.
\newblock \emph{Machine learning}, 8\penalty0 (3):\penalty0 229--256, 1992.

\end{thebibliography}

\section{Appendix}

\subsection{Implementation Details}
There are some notable differences between our implementation of APG (which we call PAGP for parallel apg) and the implementation of APG provided Brax. First, we use different controller architectures, our method uses a deterministic GRU, and theirs  uses a stochastic multi layer perceptron. Furthermore the parellization characteristics are quite different, the Brax implementation of APG was designed for use with a TPU, and thus performs hundreds or thousands of rollouts in parallel for every update step. We note that we found the performance of CE-APG to be significantly faster on CPU compared to GPU/TPU.

\subsection{Network Architecture}

Each policy consists of GRU cell with two fully connected layers on the output. The fully connection network has ReLU hidden activations and a tanh non-linearity on the final layer.

\begin{center}
\begin{tabu}{ l | l | l }
Environment & GRU Size & FC size \\
 \hline
 Cartpole Pendulum & 4  &  4x16x16x1\\ 
 Acrobot & 4  &  4x16x16x1  \\ 
 Cartpole Double Pendulum & 6 & 6x32x32x1 \\ 
\end{tabu}
\end{center}

\subsection{Hyper Paramaters}

In all cases we selected the best hyper parameters we could find using a manual search, using a coarse parameter sweep as a starting point.

\begin{center}
\begin{tabu}{ X[2,l] | X[1,l] }
CE-APG Hyperparameter & Value \\
 \hline
 APG Epochs & 100 \\ 
 Total Epochs & 200  \\ 
 Total Env Interactions &  1e7 \\
 initial std & 0.05 \\
 learning rate & 1e-3 $\rightarrow$ 1e-6 \\
 batch size (N) & 4 \\
 $K_{a}$ & 24 \\
 $K_{e}$ & 8 \\
\end{tabu}
\end{center}

\begin{center}
\begin{tabu}{ X[2,l] | X[1,l] }
PPO Hyperparameter & Value \\
 \hline
 Total Timesteps & 8e7 \\ 
 Minibatch Size & 32 \\
 Batch Size & 256 \\
 Unroll Length  & 50 \\
 N Update Epochs & 8 \\
 Discounting & 0.99 \\
 Learning Rate & 3e-4 \\
 Entropy Cost & 1e-3 \\
 N Envs & 512 \\
\end{tabu}
\end{center}

\begin{center}
\begin{tabu}{ X[2,l] | X[1,l] }
SAC Hyperparameter  & Value \\
 \hline
 Total Timesteps & 2e6 \\ 
 Discounting & .95 \\
 Learning Rate & 1e-3 \\
 N Envs & 64 \\
 Batch Size & 128 
\end{tabu}
\end{center}

\begin{center}
\begin{tabu}{ X[2,l] | X[1,l] }
Brax-APG Hyperparameter & Value \\
 \hline
 Total Env interactions & 1e7 \\ 
 N Environments & 24  \\
 learning rate & 5e-4 \\
\end{tabu}
\end{center}

\section{Environment Details}

\subsection{Acrobat}
The Acrobat is kept relatively simple, the only state variables are the joint angles and velocities, the reward is just the negative squared distance between the goal state and the current state: 

\[r_{a} = -\phi_{1}^{2} - \phi_{2}^{2} \]

With $\phi_{1} = \phi_{2} = 0$ being the upright balanced position with both links straight up.

\subsection{Cartpole Pendulum}
The cart-pole pendulum is also kept simple, the states are simply the joint angles and velocities, and the reward function is simply the negative square of the pendulum angle, with $\phi_{1}$ = 0 corresponding the the upright position.

\subsection{Inverted Double Cartpole Pendulum}

The double cartpole pendulum is modified from Brax's existing benchmark environments, the only difference is that the initial condition is rotated 180 degrees from the upright. These environments use some reward shaping, adding an an alive bonus as well as some feature extraction. rather than directly feeding in joint angles, both the reward and state variables are fed in as the x,y coordinate for the end of each link. The reward function for the environment is:

\begin{equation} 
r_{dp} = r_{alive} - r_{distance} - r_{velocity}
\end{equation}

Where 

\begin{equation}
r_{alive} = 10 
\end{equation}

\begin{equation}
r_{distance} = 0.05x^{2} + (y - y_{des})^{2}
\end{equation}

\begin{equation}
r_{velocity} = \dot \phi_{1} + \dot \phi_{2}
\end{equation}

and $y_{des}$ co-responds to the height of the second link when in the upright position. Put another way, the reward function is an alive bonus minus the euclidean distance between the end of the second link and the goal state. 






\end{document}